\documentclass[journal]{IEEEtran}
\ifCLASSINFOpdf
\else
\fi
\usepackage{amsmath}
\usepackage{amssymb}
\usepackage{booktabs}
\usepackage{tabularx}
\usepackage{CJKutf8}
\usepackage{graphicx}
\begin{document}
%
\title{Climate Change from Large Language Models}
%
%
%

\author{Hongyin~Zhu,
        Prayag~Tiwari

\thanks{H. Zhu (e-mail: hongyin\_zhu@163.com).}
\thanks{P. Tiwari is with the School of Information Technology, Halmstad University, Sweden (prayag.tiwari@ieee.org).}
}

%
%

\markboth{Journal of IEEE}%
{Shell \MakeLowercase{\textit{et al.}}: Bare Demo of IEEEtran.cls for IEEE Journals}
%



\maketitle

\begin{abstract}
Climate change poses grave challenges, demanding widespread understanding and low-carbon lifestyle awareness. Large language models (LLMs) offer a powerful tool to address this crisis, yet comprehensive evaluations of their climate-crisis knowledge are lacking. This paper proposes an automated evaluation framework to assess climate-crisis knowledge within LLMs. We adopt a hybrid approach for data acquisition, combining data synthesis and manual collection, to compile a diverse set of questions encompassing various aspects of climate change. Utilizing prompt engineering based on the compiled questions, we evaluate the model's knowledge by analyzing its generated answers. Furthermore, we introduce a comprehensive set of metrics to assess climate-crisis knowledge, encompassing indicators from 10 distinct perspectives. These metrics provide a multifaceted evaluation, enabling a nuanced understanding of the LLMs' climate crisis comprehension. The experimental results demonstrate the efficacy of our proposed method. In our evaluation utilizing diverse high-performing LLMs, we discovered that while LLMs possess considerable climate-related knowledge, there are shortcomings in terms of timeliness, indicating a need for continuous updating and refinement of their climate-related content.
\end{abstract}


%
\IEEEpeerreviewmaketitle

\section{Introduction}
The climate crisis, exacerbated by fossil fuel burning, deforestation, and industrial processes, poses a grave global threat. Its impacts range from rising sea levels to intensified weather events and biodiversity loss. Addressing this crisis is urgent, prompting widespread efforts to reduce greenhouse gas emissions and adopt more sustainable practices \cite{schimanski2024bridging}. In this context, large language models (LLMs) like GPT-4 \cite{openai2023gpt4} can play a vital role in raising awareness and educating the public about the climate emergency. LLMs have the potential to reach a global audience and provide accurate, up-to-date information on the causes and consequences of the climate crisis. They can also engage in discussions \cite{stede2021climate} with users, answering questions and addressing concerns related to climate change.

Existing LLMs have access to a significant amount of information related to the climate crisis, but this knowledge is often underutilized due to the models' lack of interpretability. Furthermore, the quality of climate crisis-related responses generated by LLMs has not been thoroughly evaluated, which limits their potential to provide valuable insights to researchers, policymakers, and other stakeholders involved in addressing climate issues. Existing methodologies for evaluating LLMs in general domains are inadequate for climate-crisis knowledge. This paper aims to analyze the challenges and opportunities associated with leveraging LLMs for climate crisis knowledge and propose a methodology to extract and assess the quality of this knowledge in an explainable way. Our approach involves eliciting climate crisis knowledge from LLMs through designed prompts and evaluating the quality of this knowledge using comprehensive metrics.

Extracting climate crisis knowledge from LLMs is a non-trivial task due to limited interpretability. Our approach aims to improve understanding and evaluation of this knowledge, enabling a more human-interpretable assessment of their capabilities. We symbolize the parameter knowledge in the text through elaborately designed prompts. To assess the knowledge accurately, we require a substantial number of relevant questions and answers. We developed a pipeline to generate and curate such questions by combining outputs from LLMs with public datasets. We then utilize LLMs to provide answers to these questions.

The second challenge is evaluating knowledge related to the climate crisis. Prior studies have primarily relied on perplexity to assess generated content, but this approach falls short in accurately capturing knowledge from a human cognitive perspective. Certain research efforts have resorted to human evaluation, an approach that can be both costly and time-consuming. Other studies have attempted to utilize classifiers to grade answers, yet these methods prove inadequate for accurately evaluating knowledge pertinent to the climate crisis. To address this issue, we propose a method to automatically evaluate the knowledge of LLMs related to the climate crisis by evaluating the quality of questions and answers. We first propose 5 metrics for evaluating questions (importance, clarity, relevance, difficulty, and innovation) and another 5 metrics for evaluating answers (relevance, depth, readability, innovation, and timeliness). We leverage high-performing LLMs to score questions and answers, then average the scores for comprehensive assessment. This integrated approach enhances evaluation accuracy and reliability.

The contributions of this paper are as follows:

(1) We propose a method to symbolize and assess the knowledge of climate crisis within LLMs.

(2) We present an approach to collect questions and answers related to the climate crisis and use LLMs to automatically evaluate the LLMs' knowledge related to the climate crisis.

(3) We introduce 5 question metrics and 5 answer metrics for objective scoring. Experimental findings validate the effectiveness of our method and highlight the limitations of LLMs in this context.

\section{Related Work}
\subsection{Large Language Models for Climate Change}
Global climate change is a significant challenge that necessitates a multidisciplinary approach. Artificial intelligence (AI) and natural language processing (NLP) technologies, such as ChatGPT, have potential applications \cite{kraus2023enhancing} in climate research, including model parameterization, data analysis, scenario generation, and evaluation. These techniques contribute to enhancing the accuracy of climate predictions and provide robust tools for researchers and policymakers.
Machine learning (ML) workloads \cite{rolnick2022tackling} are rapidly growing in importance, but their carbon footprint is a concern. Google has managed to keep ML training energy use below 15\% of total energy use over the past three years by implementing best practices. It is suggested that these practices be adopted throughout the ML field to significantly reduce the carbon footprint of model training. The application of LLM technology contributes to accurately analyzing the trends and impacts of climate change, providing strong support for sustainable development in the field of ESG (Environment, Social, and Governance) \cite{stammbach2022environmental}, and promoting the achievement of a green and low-carbon future.

LLMs, like GPT-3, are widely used in various fields, including entertainment, health, and finance \cite{zhu2023fqp}. However, their performance can be uneven when interacting with different social groups \cite{vaghefi2023chatclimate}. \cite{leippold2023thus} suggest an analytical framework to evaluate fairness in human-AI conversations. By analyzing over 20,000 conversations about climate change and the Black Lives Matter movement, they find that GPT-3 performs well when engaging with educational and minority groups regarding viewpoints. These groups not only received accurate and unbiased information but also changed their attitudes and expressed support for related actions after the dialogue.
LLMs have achieved remarkable results in AI, but they still use imprecise language in areas where accuracy is critical, such as climate change. \cite{webersinke2021climatebert} overcome its limitations and improve reliability by treating LLM as a proxy for accessing multiple sources such as ClimateWatch and general Google searches for the latest accurate climate data.

Climate change poses a significant threat to human health, and effective, evidence-based policies are needed to mitigate or eliminate these risks. This necessitates the translation of scientific knowledge into policy. To address this challenge, \cite{cheng2022gpt} propose the development of domain-specific language models for climate and health to capture available knowledge and solve various tasks, such as identifying similarities between climate and health concepts, fact-checking, extracting relationships, and generating policy text.
\cite{biswas2023potential} conducted a study on the application of ChatGPT in climate data analysis, scenario generation, and model evaluation. The research provided valuable tools for both researchers and policymakers.
\cite{patterson2022carbon} interviewed GPT-3 on the topic of climate change. Their study highlights the capabilities of LLMs but also notes that they can sometimes generate incorrect or nonsensical responses, a phenomenon known as hallucinations. The researchers will focus on strategies to prevent such hallucinations, making the models more reliable through techniques like reinforcement learning \cite{tiwari2021dapath}, and exploring the potential applications of GPT-3 in finance \cite{caldecott2022spatial,zhu2022financial} and other relevant domains.

\subsection{Large Language Models for Human Evaluation}
Large Language Models achieve controllability through human feedback mechanisms and fine-tuning the model to match human preferences. However, this approach has limitations, including complexity and instability. To address these challenges, \cite{rafailov2023direct} proposed an algorithm called Direct Preference Optimization (DPO). DPO accurately optimizes the constrained reward maximization problem in a single stage by establishing a mapping between the reward function and the optimal policy.
The application of LLMs in the medical field has sparked widespread discussion. However, they face challenges such as the potential spread of misinformation and the risk of data manipulation. \cite{reddy2023evaluating} evaluates the regulatory mechanisms that should be in place when applying LLMs to healthcare, as well as methods for assessing their performance and practical value. These efforts aim to ensure public trust in these models.
\cite{wang2023large} highlight that large language models, including GPT-4, exhibit biases in assessing the quality of responses generated by different models. By altering the sequence of responses within a context, it is possible to manipulate the evaluation outcomes to favor one model over others. To address this issue, they developed a calibration framework that incorporates three straightforward and effective strategies: multi-evidence calibration, balanced position calibration, and human cycle calibration. These methods help to reduce evaluation bias and align the results more closely with human judgment.

KoLA \cite{yu2023kola} is a meticulously crafted knowledge-centric evaluation benchmark designed to assess the capabilities of LLMs. The benchmark features a four-tiered classification system for knowledge-related abilities \cite{zhu2023pre}, which emulates human cognition. Additionally, it incorporates data from Wikipedia and other sources that are regularly updated. KoLA employs an evaluation methodology that utilizes both standard scores and self-comparison indicators. The authors evaluated 21 open-source and commercial LLMs and conducted a thorough analysis of their findings.
\cite{DBLP:conf/acl/ChiangL23} investigated whether large language models could serve as a substitute for human evaluation. The study compared the use of LLMs and human evaluators in assessing text quality for two natural language processing tasks. The findings indicate that the evaluation outcomes generated by LLMs align with those provided by human experts. The researchers discovered that the results from LLM evaluations remained consistent across different formats of task instructions and were deemed stable and reliable. The paper further discusses the limitations and ethical implications of using LLMs for assessment purposes.

\section{Approach}
\begin{figure}[!h]
\centering
\includegraphics[width=\linewidth]{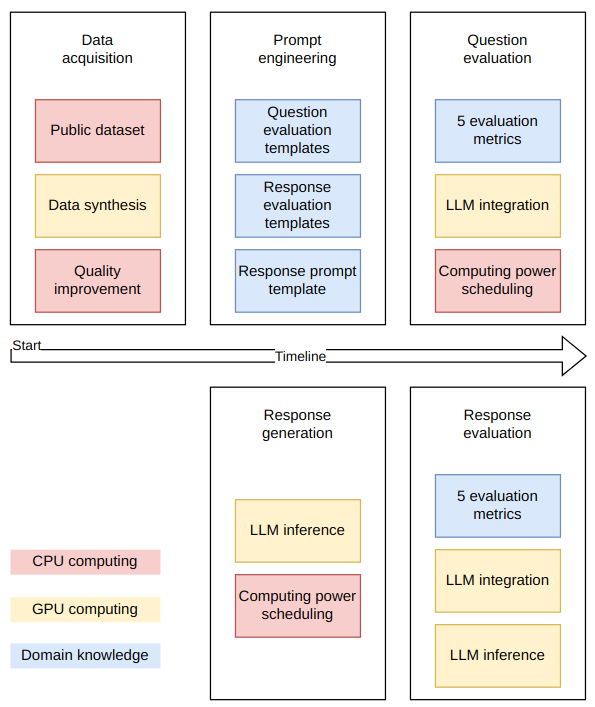}
\caption{The schematic diagram of the proposed climate crisis knowledge evaluation framework}
\label{arch2.fig}
\end{figure}

We formalize the climate crisis knowledge evaluation task. Given a set of climate crisis questions $X^{(q)}=\{x_1^{(q)},x_2^{(q)},... \}$ and answers $X^{(a)}=\{x_1^{(a)},x_2^{(a)},... \}$, we use LLMs as evaluators to generate responses based on predefined metrics, which reflect the knowledge contained within the model. Unlike previous work, the innovation of this paper is that we propose an automatic LLM inference framework that evaluates the climate-crisis knowledge of LLMs from 10 different perspectives. The overview framework is shown in Figure \ref{arch2.fig}. The timeline includes data acquisition, prompt engineering, question evaluation, response generation, and response evaluation. These modules can be processed in parallel. In this section, we first introduce the acquisition of Climate Crisis Questions and Answers, followed by an introduction to the Climate Crisis Knowledge Evaluation.

\subsection{Acquisition of Climate-Crisis Q\&A Dataset}
Our proposed method for acquiring a large number of questions about the climate crisis involves a two-step process: question generation and question selection. Initially, we used the Llama2-70B \cite{touvron2023llama} model to generate 100,000 questions. This model has advanced language understanding and generation capabilities, enabling the creation of a diverse range of questions that cover various aspects of the climate crisis. After generating the questions, we perform a thorough classification and labeling process to facilitate efficient analysis and processing of the questions.

Following our initial selection, we conducted an additional review to eliminate questions that were irrelevant or duplicates. We established a set of rules to guide this process, which was fully automated with no human intervention. To ensure the quality of the questions, we improved their quality through the following steps: (1) Removal of overlapping questions: Through semantic analysis, we identified redundant questions and employed an embedding-based question retrieval method to retain only unique questions, effectively eliminating duplicates based on a defined threshold, as shown in equation \eqref{alpha}. (2) Climate crisis relevance assessment: We conducted a relevance analysis of each question to ensure its practical significance and value in the context of the climate crisis, leveraging keyword occurrence.
\begin{align}
\alpha_{i,j} = \frac{{h_i^{(q)}}^T h_j^{(q)}}{||h_i^{(q)}||\cdot||h_i^{(q)}||}\qquad i,j \in m, i\neq j
\label{alpha}
\end{align}
where $h_i^{(q)} \in \mathbb{R}^d$ is determined using equation \eqref{encoder}.
\begin{align}
h_i^{(q)} = \mathcal{F}_{encoder}(x_i^{(q)}|\Theta)
\label{encoder}
\end{align}
where $\mathcal{F}_{encoder}(\cdot)$ is is a language model for generating embeddings \cite{bge_embedding}. $\Theta$ represents the parameters of the model. $x_i^{(q)}$ is the sequence of text in the question. 

After processing these questions, we obtained a valuable collection of 19,241 high-quality questions related to the climate crisis, about 5\% of this data came from an external dataset. Since LLMs are pre-trained with the next token prediction task, as shown in equation \eqref{gpt}, we subsequently leveraged Llama2-70B to generate corresponding answers for each question \cite{zhu2021collaborative}. Our two-stage methodology effectively facilitated the accumulation of a substantial number of question-answer pairs.
\begin{align}
p(x) = \prod_{i=1}^n p(w_i|w_{i-1},...,w_1,\Theta)
\label{gpt}
\end{align}
where $x$ is the input text and $w_i$ represents the $i$-th token. $\Theta$ is the model parameter. In the following, we introduce a novel methodology for assessing knowledge about the climate crisis. Our approach aims to establish an objective and precise criterion for evaluating questions and answers related to this critical topic, leveraging the capabilities of multiple LLMs.

\subsection{Evaluation of Climate-Crisis Knowledge}
\begin{figure}[!htbp]
\centering
\includegraphics[width=0.8\linewidth]{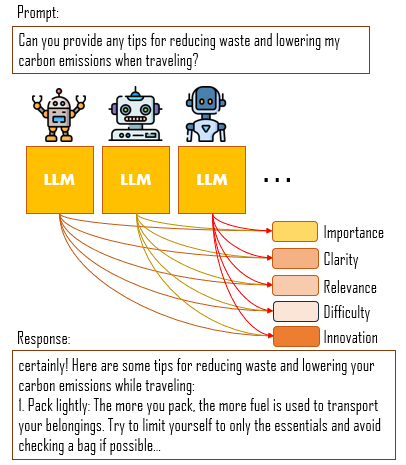}
\caption{An illustration of utilizing multiple LLMs to automatically evaluate a question-answer pair in the context of climate change}
\label{arch.fig}
\end{figure}
We use multiple LLMs to generate scores for the questions, as shown in Figure \ref{arch.fig}. To allow the model to evaluate the responses from various aspects, we developed several prompt templates \cite{zhu2023metaaid} for questions and answers, including different types of questions, so that the model can be evaluated from multiple perspectives. For instance, the prompt might be: "Please assess the importance of the above question: How valuable is this question to the user? Can it help users express their needs and confusion?" or "Please rate the clarity of the above questions: Is the question clear and easy to understand?" In this way, the model can rate each question and answer based on its learned knowledge.

To evaluate the quality of the questions, we evaluate them from the following aspects: (1) Importance of the problem: How valuable is this problem to the user? Can it help users express their needs and confusion? (2) Clarity of the question: Is the question clear and understandable at a glance? (3) Relevance of the question: Is the question closely related to the topic? (4) Question difficulty: Is the question too difficult or too easy for users to understand or too simple to interest users? (5) Innovation of the question: Is the question novel and can it inspire users to think?

To evaluate the quality of answers, we evaluate the following aspects: (1) Relevance of the answer: Does the answer accurately answer the user's question and can it solve the user's needs? (2) Depth of answer: Does the answer provide enough detail so that users can fully understand and apply the information? (3) Answer readability: Is the answer written in plain language and clearly formatted for users to read and understand? (4) Innovation of the answer: Does the answer provide unique insights or solutions that will help users achieve better results on similar problems? (5) Timeliness of the answer: Is the content of the answer up-to-date and able to adapt to changing circumstances and needs?

We use the model to automatically score the metrics mentioned above. For the question, we use equation \eqref{questemp}. 
\begin{align}
x_{i,j}^{(r)} = \mathcal{F}_{decoder}(<x_i^{(q)}; [pre]^{(q)}; m_j^{(q)}; [suf]^{(q)}>)
\label{questemp}
\end{align}
where the prefix and suffix of the template are denoted as $[pre]^{(q)}$ and $[suf]^{(q)}$, respectively. The $j$-th metric for question evaluation is represented as $m_j^{(q)}$. The LLM is denoted as $\mathcal{F}_{decoder}(\cdot)$. The concatenation operation is represented as $<;>$.

For the answer, we use equation \eqref{anstemp}. Then we can get the generation content as the candidate data. Finally, we extract the model scores from the data using information extraction methods.
\begin{align}
x_{i,j}^{(r)} = \mathcal{F}_{decoder}(<x_i^{(q)}; x_i^{(a)}; [pre]^{(a)}; m_j^{(a)}; [suf]^{(a)}>)
\label{anstemp}
\end{align}
where the prefix and suffix of the template are denoted as $[pre]^{(a)}$ and $[suf]^{(a)}$, respectively. The $j$-th metric for answer evaluation is represented as $m^{(a)}$.

Then we manually check the model scores through random sampling. We found that the model's evaluation of the quality of generated responses is highly consistent with that of humans. We also discovered some potential problems. In some cases, the model may misunderstand the intent of the instruction, resulting in an invalid response. Additionally, because the model is trained on a massive amount of cross-domain text data, it may not fully understand certain aspects of the climate crisis or questions of a metaphorical nature. To address these issues, we can further fine-tune the LLMs in the future to improve their ability to understand complex questions and answers.

We anticipate that this methodology will foster a deeper comprehension of climate crisis-related issues among individuals and offer a fair and unbiased evaluation criterion. In practical scenarios \cite{zhu2022metaaid}, when users submit questions or answers, LLMs will seamlessly process them and assign a corresponding score, without human intervention, using predefined prompt templates. This not only empowers the system to deliver insightful answers but also assesses the quality of the information, ultimately assisting users in grasping the gravity of climate crisis-related topics. By employing carefully crafted prompt templates, our approach guarantees an objective and precise evaluation of climate crisis-related questions and answers, thus contributing significantly to heightening public awareness and encouraging greater participation in the fight against climate change.

\section{Experiments}
\subsection{Dataset}
We curated a comprehensive climate-crisis Q\&A dataset that encompasses a vast array of questions and answers about climate change. This dataset boasts a total of 19,241 samples, of which 95\% of the questions were intelligently generated using the Llama2-70B model. The remaining 5\% of questions were carefully sourced from pertinent information gathered from the internet, including the "Reddit Climate Change Dataset" that captures discussions about climate change on Reddit up to September 1, 2022. This dataset comprises 620,908 posts and 4,600,698 comments. To ensure the quality and relevance of our dataset, we employed a rigorous two-step processing method to eliminate any overlapping content and enhance its relevance. The answers within this dataset are also automatically generated by the Llama2-70B model.

\subsection{Hyper-parameters}
We employ several high-performing LLMs for evaluation, with the temperature parameter set to 0.5 for all models, and a maximum length of 2048. We set do\_sample to false to ensure reproducibility of results. For Llama2-70b, we use top\_k = 250, top\_p = 1, and repetition\_penalty = 1. For Baichuan2-13b, we set top\_k = 5, top\_p = 0.85, and repetition\_penalty = 1.05. For the remaining models, we adhere to their default configurations. The experimental environment consists of an Intel(R) Xeon(R) Platinum 8163 CPU @ 2.50GHz with 256G of memory, and 8 RTX 3090 (24G) GPUs.

\subsection{Evaluation}
We assess the quality of questions and answers utilizing LLMs, which assign a rating on a scale ranging from 0 to 10. A higher score signifies superior quality in each respective aspect. Specifically, we employ 5 distinct metrics to evaluate questions and another set of 5 metrics to evaluate answers.

\subsection{Question Quality Evaluation}
%
The following models were adopted to assess the quality of questions.

The ChatGLM3-6B model \cite{zeng2022glm} is built upon the GLM architecture and employs an autoregressive blank infilling training approach. This training method equips the model to handle complex scenarios with ease, including tasks like tool invocation (function call), code execution (code interpreter), and Agent-related tasks.

Mistral-7B \cite{jiang2023mistral} uses grouped-query attention and sliding-window attention, and it employs a byte-fallback BPE tokenizer. The model is designed to strike a balance between efficiency and performance by creating a lightweight architecture.

Zephyr-7B \cite{tunstall2023zephyr} is based on the Mistral-7B model and employs the direct preference optimization (DPO) training approach.

The Baichuan2-13B model \cite{yang2023baichuan} is trained on a high-quality corpus of 2.6 trillion tokens. The model employs ALiBi linear bias technology to enhance computational efficiency effectively.

The Yi-34B model \cite{Yi2023} is based on the Llama model architecture and has been trained on both Chinese and English data. It uses approximately 3T-sized tokens and supports long-context technology.

Llama2-70B \cite{touvron2023llama}, as proposed by Meta, is an open-source model architecture that has been trained using reinforcement learning with human feedback (RLHF). This training methodology is designed to align the model's behavior with human preferences, ensuring both its usefulness and safety.

\begin{table*}[htbp!]
\centering
\caption{Question quality evaluation scores (0-10) across 5 dimensions}
\begin{tabular}{c|ccccc} \toprule
Models   & Importance &  Clarity   &  Relevance &   Difficulty & Innovation \\ \midrule
ChatGLM3-6B   &  8.37  & 8.47  & 8.56   &  6.27  &  8.28 \\
Mistral-7B  & 8.25  & 7.80 & 9.06 & 6.77  & 8.36 \\
Zephyr-7B &  9.95 & 9.98 & 9.97  & 6.45  & 9.45 \\
Baichuan2-13B   & 8.70  &  9.00   & 8.37  & 7.60 & 8.79  \\
Yi-34B   & 8.84  & 8.64 & 9.75  & 6.78  & 7.18 \\
Llama2-70B  & 8.72 &  8.33   & 8.79  & 7.15  & 8.00  \\ \midrule
Average & 8.81 &   8.85  & 9.13  & 6.84  & 8.34 \\
\bottomrule       
\end{tabular}
\label{quesquality}
\end{table*}

\begin{figure}[htbp!]
\centering
\includegraphics[width=\linewidth]{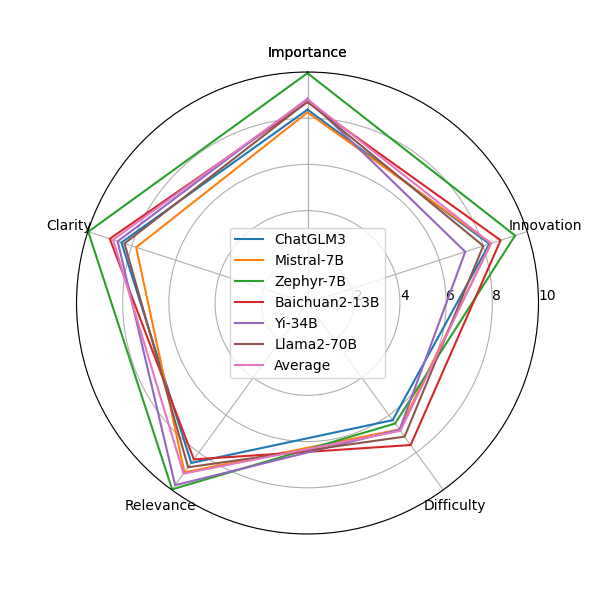}
\caption{Visualization of question quality evaluation, with circles closer to the center indicating lower overall scores assigned by the model}
\label{question.fig}
\end{figure}
As can be seen from the results in Table \ref{quesquality}, the model evaluation results suggest that the quality of questions is highest concerning topic relevance but lowest concerning question difficulty. This means that integrating multiple LLMs can produce more credible results for climate-crisis knowledge.

As shown in Figure \ref{question.fig}, the curve closer to the outside indicates a higher overall score for the model. Zephyr-7B gives the highest overall score of question quality, while Mistral-7B and Yi-34B give the lower overall score of question quality. This means that different models have different standards for knowledge about the climate crisis, and we tend to choose models with more stringent standards.
\subsection{Answer Quality Evaluation }
\begin{table*}[htbp!]
\centering
\caption{Answer quality evaluation scores (0-10) across 5 dimensions}
\begin{tabular}{c|ccccc} \toprule
Models   & Relevance &  Depth   &  Readability &   Innovation & Timeliness \\ \midrule
ChatGLM3-6B   & 9.92  & 8.42  &  8.82 & 8.73 & 8.39  \\
Mistral-7B   &  8.67 &  8.98 & 9.00  &  8.55 & 8.22 \\
Zephyr-7B   &  9.98 & 9.89  & 9.95  & 9.75  & 9.95 \\
Baichuan2-13B   & 8.73  & 8.41  & 8.48  &  8.23 & 6.65 \\
Yi-34B   & 9.65  & 9.00  &  9.22 &  7.54 & 8.17 \\
Llama2-70B   & 9.11 &  9.10  &  9.35 & 8.97  & 9.12  \\ \midrule
Average   &  9.34 & 8.97  &  9.14 & 8.63  & 8.42 \\
\bottomrule       
\end{tabular}

\label{ansquality}
\end{table*}
As can be seen from the results in Table \ref{ansquality}, the model evaluation results suggest that the quality of the answers is the highest in terms of relevance, but is low in terms of question timeliness. This means that LLMs can understand climate crisis knowledge and are accustomed to generating relevant responses, but contain insufficient timely information.

\begin{figure}[htbp!]
\centering
\includegraphics[width=\linewidth]{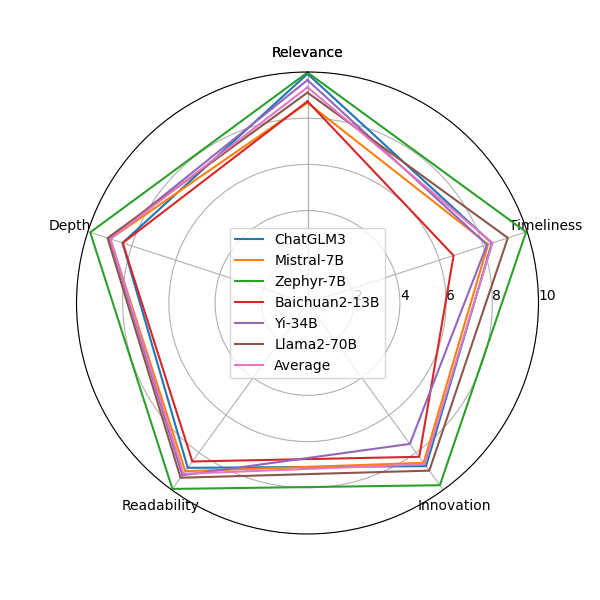}
\caption{Visualization of answer quality evaluation, with circles positioned closer to the center indicating lower overall scores assigned by the model}
\label{answer.fig}
\end{figure}

As shown in Figure \ref{answer.fig}, the curve closer to the outside indicates a higher overall score for the model. Among the models evaluated, Zephyr-7B gives the highest overall score for answer quality, while Baichuan2-13B gives a lower overall score for answer quality. We can find that different models have different sensitivity to the timeliness of answers.

\subsection{Computing Efficiency Analysis}
We compare LLMs and conduct experiments using 4-bit quantization to ensure optimal efficiency. We evaluate model performance using 10 different prompts and set a maximum sequence length of 2048.

As shown in Table \ref{computtab}, ChatGLM3-6B has the fastest inference speed, while Llama2-70B has the lowest speed. Due to the high GPU memory needs, we averagely split different layers of Llama2-70B to 8 GPUs, so it has extra time consumption to communicate among PCIe GPUs. Mistral-7B and Zephyr-7B perform poorly compared to similarly sized models. For the GPU memory consumption, we found that Llama2-70B has the best GPU memory utilization efficiency (0.51GB/1B). ChatGLM3-6B has the worst memory utilization efficiency (0.80GB/1B).

\begin{table}[htbp!]
\centering
\caption{Efficiency of Inference for Various LLMs}
\begin{tabular}{c|cc} \toprule
Models   & Time (S) &  Memory (GB)  \\ \midrule
ChatGLM3-6B   & 26.66  &  4.80 \\
Mistral-7B   & 179.09  &  5.37 \\
Zephyr-7B   &  325.14 &  5.33 \\
Baichuan2-13B   & 44.12  & 11.79   \\
Yi-34B   &  70.39 &  19.48  \\
Llama2-70B   & 709.63 &  36.20   \\ 
\bottomrule       
\end{tabular}
\label{computtab}
\end{table}

\subsection{Case Study}

We use the question "Can you provide any tips for reducing waste and lowering my carbon emissions when traveling?" as an example to assess the responses of the evaluation models.

As shown in Table \ref{comparison} located in the Appendix, each of the models adopted is capable of providing high-quality responses. Among them, Yi-34B stands out for offering the most comprehensive suggestions, while Llama2-70B is known for providing a response that is both concise and effective.

\section{Conclusion}

This paper introduces an automated framework for evaluating the climate-crisis knowledge of LLMs. Our proposed approach assesses climate-crisis knowledge based on the quality of symbolized questions and their corresponding answers. The evaluation process is crafted to be both robust and comprehensive, encompassing a two-stage question acquisition strategy and an answer generation procedure. Furthermore, we have devised an automated evaluation methodology along with a comprehensive set of metrics, including 5 for question evaluation and 5 for answer evaluation. Experimental findings indicate that our approach holds significant value in assessing LLMs' knowledge pertaining to climate change.

The primary contribution of this paper is the proposal of an automated framework to evaluate climate-crisis knowledge in LLMs, without reliance on human intervention. Looking forward, we aim to leverage this technique in the development of an online climate crisis knowledge system that utilizes our methodologies to provide users with real-time, expert-level Q\&A services. Our research introduces novel concepts and methodologies that address challenges in the field of climate crisis, thereby enriching the research and applications of AI in this critical domain.

\bibliographystyle{IEEEtran}
\bibliography{reference}

\appendix

\section{Appendix}
\label{sec:appendix}

\begin{CJK*}{UTF8}{gbsn}
\begin{table*}[]
\centering
\tiny
\caption{Comparison of responses from different models}
\begin{tabularx}{\linewidth}{c|X } \toprule
Models   & \multicolumn{1}{c}{Response}   
\\ \midrule
ChatGLM3-6B   & 当然可以，以下是一些减少浪费和降低碳排放的旅行建议：\newline
1. 选择环保的交通工具：选择公共交通工具、自行车或步行，而不是自驾或飞机。火车和巴士通常比飞机更环保，尤其是长途旅行。\newline
2. 飞行时选择直飞航班：直飞航班通常比转机航班更环保，因为它们减少了燃料消耗和碳排放。\newline
3. 优化旅行计划：提前规划行程，避免不必要的绕路和重复旅行，减少碳足迹。\newline
4. 住宿选择：选择环保型酒店或民宿，这些地方通常采取措施减少能源消耗和浪费。\newline
5. 自带餐食：避免经常在外就餐，而是自带食物和饮料，这样可以减少一次性餐具的使用。\newline
6. 减少一次性用品的使用：尽量避免使用一次性物品，如塑料袋、瓶子和餐具。携带可重复使用的物品，如水壶、餐具和购物袋。\newline
7. 回收和节约资源：在住宿和旅行时，尽量回收利用废弃物，并节约用水和用电。\newline
8. 支持环保旅行：选择参加环保旅行团或活动，这些活动通常会考虑对环境的影响。\newline
9. 购买碳抵消：如果必须乘坐飞机等高排放交通工具，可以考虑购买碳抵消，以补偿碳排放。\newline
10. 分享行程：如果可能，与其他人分享行程，这样可以减少整体碳足迹。\newline
记住，每次旅行都会对环境产生影响，但通过采取上述措施，您可以减少对环境的影响，并帮助保护我们的地球。   \\ \hline
Mistral-7B   &  Sure, here are some tips for reducing waste and lowering your carbon emissions when traveling:\newline
1. Pack light: Try to pack only what you need for your trip. This will reduce the amount of weight you need to carry, which will in turn reduce your carbon emissions.\newline
2. Use reusable containers: Instead of using disposable plastic bags or containers, bring your own reusable containers for food and other items. This will reduce waste and save money.\newline
3. Bring your own water bottle: Instead of buying bottled water, bring your own reusable water bottle. This will reduce waste and save money.\newline
4. Use public transportation: Instead of renting a car or taking taxis, use public transportation to get around. This will reduce your carbon emissions and save money.\newline
5. Stay in eco-friendly accommodations: Look for accommodations that are eco-friendly, such as hotels that use renewable energy sources or have recycling programs.\newline
6. Eat locally: Instead of eating at chain restaurants, try to eat locally. This will reduce your carbon emissions and support the local economy.\newline
7. Use reusable bags: Instead of using plastic bags, bring your own reusable bags for shopping and other activities. This will reduce waste and save money.\newline
8. Use energy-efficient electronics: Bring energy-efficient electronics, such as a portable charger and a laptop with a long battery life. This will reduce your energy consumption and save money.\newline
By following these tips, you can reduce your waste and lower your carbon emissions when traveling.
 \\ \hline
Zephyr-7B   &  Certainly! Here are some tips for reducing waste and lowering your carbon emissions when traveling:\newline
1. Pack light: The more you pack, the more fuel is needed to transport your luggage. Pack only the essentials and avoid overpacking.\newline
2. Choose eco-friendly accommodations: Look for accommodations that prioritize sustainability, such as those with green certifications, solar panels, or recycling programs.\newline
3. Use reusable water bottles and bags: Bring a reusable water bottle and bag to avoid using single-use plastics. Many destinations have water refill stations, so you can fill up your bottle for free.\newline
4. Eat local and seasonal: Eating locally sourced and seasonal foods reduces the carbon footprint of transportation and supports the local economy.\newline
5. Use public transportation: Use public transportation, such as buses, trains, and subways, instead of renting a car. This reduces your carbon emissions and saves you money.\newline
6. Offset your carbon emissions: Consider offsetting your carbon emissions by investing in carbon-reducing projects, such as reforestation or renewable energy.\newline
7. Support sustainable tourism: Choose activities that prioritize sustainability, such as nature walks, bike tours, and cultural experiences.\newline
8. Reduce water usage: Turn off the tap when brushing your teeth, take shorter showers, and reuse towels to reduce water usage.\newline
9. Support local businesses: Support local businesses, such as small shops and restaurants, to reduce the carbon footprint of transportation and support the local economy.\newline
10. Educate yourself: Learn about the local culture and environment to better understand the impact of your actions and make informed decisions.\newline
Remember, every small action can make a big difference in reducing waste and lowering your carbon emissions when traveling.
 \\ \hline
Baichuan2-13B   &  Sure, here are some tips to reduce waste and lower your carbon emissions when traveling:\newline
1. Use public transportation or carpooling instead of renting a car. This will not only save you money but also reduce your carbon footprint.\newline
2. Bring reusable water bottles, coffee mugs, and shopping bags. Avoid using single-use plastic items as much as possible.\newline
3. Book accommodation that has eco-friendly practices like energy efficiency and water conservation.\newline
4. Eat locally sourced food while traveling. This reduces the need for transportation and minimizes waste.\newline
5. Choose accommodations with energy-efficient lighting and appliances. This will help in reducing energy consumption.\newline
6. Minimize your use of air conditioning and heating. Wear layers of clothing to regulate your body temperature.\newline
7. Unplug chargers and devices not in use. This will help in reducing energy consumption.\newline
8. Consider staying at an eco-friendly hotel or resort that uses renewable energy sources.\newline
9. Use energy-efficient devices like LED lights and energy-saving appliances.\newline
10. Support local businesses and cultural activities. This will help in promoting sustainable tourism.    \\ \hline
Yi-34B   &  Yes, there are several tips to reduce waste and lower carbon emissions when traveling:\newline
1. Plan Your Trip Carefully:\newline
    Choose modes of transportation that have lower carbon emissions, such as trains or buses over flying, especially for shorter distances.\newline
    Combine multiple destinations into one trip to reduce the number of flights taken.\newline
2. Use Public Transportation:\newline
    Opt for local buses, trains, or subways instead of renting a car or using taxis.\newline
3. Ride Sharing:\newline
    Use carpooling or ridesharing services to reduce the number of vehicles on the road.\newline
4. Eco-Friendly Accommodations:\newline
    Choose hotels or hostels that have eco-friendly practices, such as energy-efficient lighting, water-saving devices, and recycling programs.\newline
5. Bring Your Own Gear:\newline
    Bring your own reusable water bottle, coffee mug, and cutlery to reduce the use of single-use items.\newline
6. Pack Lightly:\newline
    Travel with only what you need to reduce the weight and volume of your luggage, which can lower fuel consumption for planes and vehicles.\newline
7. Offset Your Carbon Footprint:\newline
    Consider investing in carbon offsets to compensate for the emissions you produce during your trip.\newline
8. Eat Locally and Seasonally:\newline
    Support local farmers and reduce the carbon footprint associated with food transportation by eating locally sourced and seasonal produce.\newline
9. Shop Wisely:\newline
    Avoid buying items that require a lot of packaging or that will end up in the trash.\newline
10. Leave No Trace:\newline
    If you're camping or hiking, follow the leave-no-trace principles, which include packing out all your trash and being mindful of the environment.\newline
11. Travel Off-Peak:\newline
    Avoid traveling during peak seasons when possible, as fewer travelers mean less environmental impact.\newline
12. Be Mindful of Water Usage:\newline
    Limit your water usage, especially in places where water is scarce.\newline
By following these tips, you can help minimize your environmental impact while traveling and contribute to a more sustainable tourism industry. Remember that every little bit helps, and by making small changes, you can make a big difference.
   \\ \hline
Llama2-70B   & certainly! Here are some tips for reducing waste and lowering your carbon emissions while traveling:\newline
1. Pack lightly: The more you pack, the more fuel is used to transport your belongings. Try to limit yourself to only the essentials and avoid checking a bag if possible.\newline
2. Use public transportation: Instead of renting a car or taking taxis, use public transportation to get around. Not only is it more environmentally friendly, but it's also a great way to experience the local culture.\newline
3. Bring a reusable water bottle: Instead of buying bottled water, bring a reusable water bottle and fill it up at a water fountain or filtered water station.\newline
4. Avoid single-use plastics: Refuse single-use plastics like straws, bags, and utensils. Instead, opt for reusable alternatives like metal straws, cloth bags, and bamboo utensils.\newline
5. Choose eco-friendly accommodations: Look for hotels, hostels, and vacation rentals that have implemented sustainability practices such as using solar power, reducing water waste, and recycling.\newline
6. Reduce food waste: Plan your meals in advance and try to avoid ordering too much food. Also, consider choosing local, seasonal ingredients which have a lower carbon footprint.\newline
7. Offset your carbon emissions: Consider offsetting your carbon emissions from your flight by investing in carbon offset projects that reduce greenhouse gas emissions elsewhere, such as reforestation or renewable energy projects.\newline
I hope these tips help you reduce your impact on the environment while traveling!      \\
\bottomrule       
\end{tabularx}
\label{comparison}
\end{table*}
\end{CJK*}
\end{document}